\documentclass{article}
\usepackage{spconf,amsmath,graphicx}

\usepackage{enumitem}
\setlist{nosep, leftmargin=14pt}

\usepackage{mwe} 

\usepackage[utf8]{inputenc} 
\usepackage[T1]{fontenc}    
\usepackage{hyperref}       
\usepackage{url}            
\usepackage{booktabs}       
\usepackage{amsfonts}       
\usepackage{nicefrac}       
\usepackage{microtype}      
\usepackage[table]{xcolor}

\usepackage{xcolor}         
\usepackage{graphicx}
\usepackage{multirow}
\usepackage{array}
\usepackage{amssymb}
\usepackage{amsmath}
\usepackage{caption}
\usepackage{bbding}
\usepackage{subcaption}
\usepackage{blindtext}
\usepackage{adjustbox}
\usepackage{amsmath}

\usepackage{graphicx}

\include{math_command}

\usepackage{amssymb}
\usepackage{pifont}

\newcolumntype{L}[1]{>{\raggedright\let\newline\\\arraybackslash\hspace{0pt}}m{#1}}
\newcolumntype{C}[1]{>{\centering\let\newline\\\arraybackslash\hspace{0pt}}m{#1}}
\newcolumntype{R}[1]{>{\raggedleft\let\newline\\\arraybackslash\hspace{0pt}}m{#1}}
\usepackage[utf8]{inputenc}
\usepackage[english]{babel}
\usepackage{amsthm}
\usepackage[misc]{ifsym}

\newcommand{\app}{\raise.17ex\hbox{$\scriptstyle\sim$}}
\makeatletter\renewcommand\paragraph{\@startsection{paragraph}{4}{\z@}
  {.495em \@plus1ex \@minus.2ex}{-.5em}{\normalfont\normalsize\bfseries}}\makeatother

\makeatletter
\DeclareRobustCommand\onedot{\futurelet\@let@token\@onedot}
\def\@onedot{\ifx\@let@token.\else.\null\fi\xspace}

\newcommand{\eg}{\mbox{e.g.,\ }}
\newcommand{\etal}{\mbox{et al.}}
\newcommand{\ie}{\mbox{i.e.,\ }}

\makeatother

\newcolumntype{x}[1]{>{\centering\arraybackslash}p{#1pt}}
\newcolumntype{a}[1]{>{\columncolor{verylightgray}\centering\arraybackslash}p{#1pt}}
\newcolumntype{y}[1]{>{\raggedright\arraybackslash}p{#1pt}}
\newcolumntype{z}[1]{>{\raggedleft\arraybackslash}p{#1pt}}\newlength\savewidth\newcommand\shline{\noalign{\global\savewidth\arrayrulewidth\global\arrayrulewidth 1pt}\hline\noalign{\global\arrayrulewidth\savewidth}}

\newcolumntype{P}[1]{>{\centering\arraybackslash}p{#1}}

\usepackage{bbding}
\usepackage{bibunits}
\usepackage{bm}
\usepackage{array}
\usepackage{xcolor}
\usepackage{pdflscape}
\usepackage{makecell}
\usepackage{framed,multirow}
\usepackage[flushleft]{threeparttable}
\usepackage{amssymb}
\usepackage{pifont}
\usepackage{algorithm}
\usepackage{layouts}
\usepackage{listings}
\usepackage{pythonhighlight}
\makeatletter
\newcommand\footnoteref[1]{\protected@xdef\@thefnmark{\ref{#1}}\@footnotemark}
\makeatother

\definecolor{Highlight}{HTML}{39b54a}  

\usepackage{xcolor}

\definecolor{amber}{rgb}{1.0, 0.49, 0.0}

\title{Label-Assemble:\\Leveraging Multiple Datasets with Partial Labels}
\name{Mintong~Kang$^{1}$, Bowen~Li$^{2}$, Zengle~Zhu$^{3}$, Yongyi~Lu$^{2}$, Elliot~K.~Fishman$^{4}$, Alan~Yuille$^{2}$, Zongwei~Zhou$^{2,*}$}
\address{$^{1}$University of Illinois Urbana-Champaign \quad  $^{2}$Johns Hopkins University\\$^{3}$Tongji University \quad $^{4}$Johns Hopkins University School of Medicine}

\begin{document}
%
\maketitle
\begin{abstract}
The success of deep learning relies heavily on large labeled datasets, but we often only have access to several small datasets associated with partial labels.
To address this problem, we propose a new initiative, ``Label-Assemble'', that aims to unleash the full potential of partial labels from an assembly of public datasets.
We discovered that learning from negative examples facilitates both computer-aided disease diagnosis and detection.
This discovery will be particularly crucial in novel disease diagnosis, where positive examples are hard to collect, yet negative examples are relatively easier to assemble.
For example, assembling existing labels from NIH ChestX-ray14 (available since 2017) significantly improves the accuracy of COVID-19 diagnosis from 96.3\% to 99.3\%.
In addition to diagnosis, assembling labels can also improve disease detection, \eg the detection of pancreatic ductal adenocarcinoma (PDAC) can greatly benefit from leveraging the labels of Cysts and PanNets (two other types of pancreatic abnormalities), increasing sensitivity from 52.1\% to 84.0\% while maintaining a high specificity of 98.0\%.
Code is available \href{https://github.com/MrGiovanni/LabelAssemble}{here}.
\end{abstract}
\section{Introduction}
\label{sec:intro}

\begin{figure*}[t]
\footnotesize
    \centering
    \includegraphics[width=1.0\linewidth]{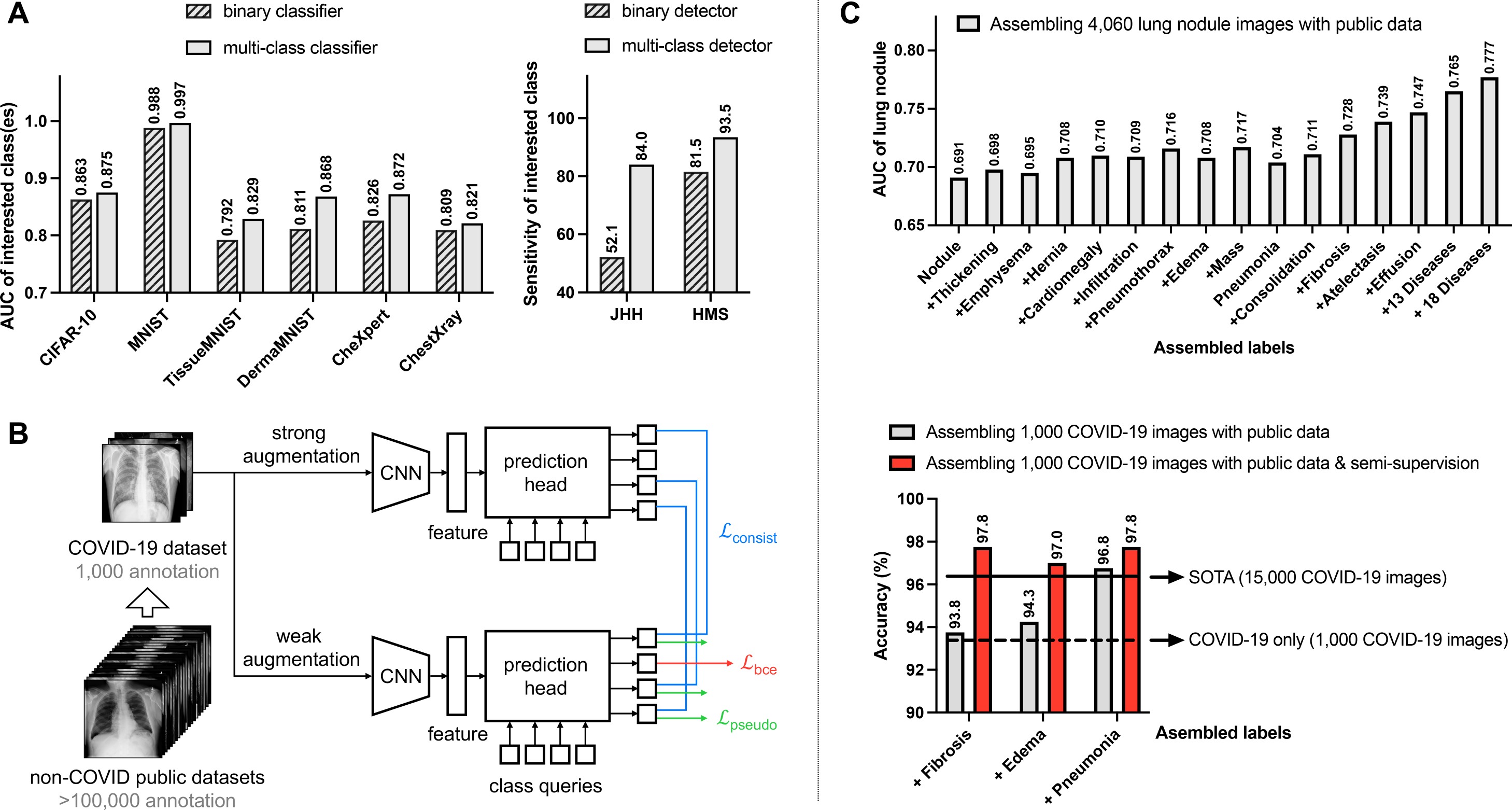}
    \caption{
    \textbf{Overview.} Our proposed framework is capable of harnessing partially labeled and unlabeled data from heterogeneous sources (\eg COVID-19 and non-COVID public datasets).
     \textbf{A.} With the same amount of data, learning from classes of ``negative examples'' benefits the learning of the interested class (see~\S\ref{sec:motivation}). 
     This observation is verified by six classification tasks and two detection tasks, serving as the foundation of the Label-Assemble initiative. 
    \textbf{B.} Labels in an assembly of public datasets are incomparable and conflicting---negative examples in the COVID-19 dataset can include the positive class in other datasets.
    A shared CNN extracts image features, and then a prediction head generates the predictions by inner producting features and class queries.
    A supervised loss ($\mathcal{L}_{\text{bce}}$) is used if the label is given; two unsupervised losses ($\mathcal{L}_{\text{pseudo}}$ \& $\mathcal{L}_{\text{consist}}$) are used if the label is absent.
    \textbf{C.} Assembling labels of other chest diseases improves lung nodule classification.
    The performance gain is \textit{positively} correlated to inter-class similarity between nodule and the assembled disease (see~\S\ref{sec:improved_diagnosis}).
    The Pearson Coefficient is $r=0.83$; $p=4.93e$-$4$. Assembling 1,000 labeled COVID-19 images with public data (available since 2017), we achieve significantly higher performance than the previous state-of-the-art method, which required over 15,000 labeled COVID-19 images. Again, pathologically similar diseases (\ie Pneumonia) lead to greater improvement in computer-aided diagnosis of COVID-19.
    }
    \label{fig:framework_overview}

\end{figure*}

Recent years have witnessed an increasing number of datasets becoming publicly available thanks to the collective efforts of imaging data archives~\cite{yang2021radimagenet} and international competitions~\cite{antonelli2021medical,baid2021rsna}.
These datasets are collected, organized, annotated differently, and often come with partial labels. 
Very few studies have been done to unleash the full potential of an assembly of multiple datasets with partial labels. 
The challenge is that labels in those public datasets are often incomparable, heterogeneous, or even conflicting~\cite{zhou2019prior,yan2020learning}. 
In this paper, we ponder the question: \emph{Can we integrate and exploit such a great number of publicly available datasets with partial labels to achieve an improved computer-aided diagnosis and detection of specific diseases?}

To address this question, we start by probing a principal hypothesis (see~\S\ref{sec:motivation}): 
\textit{a dataset that is labeled with various classes can foster more powerful models than one that is only labeled with the class of interest.}
Consequently, we propose a new initiative of ``Label-Assemble'' for leveraging partial labels from an assembly of data on hand.
Specifically, we develop a new \textit{class query} to encode different visual tasks, which can dynamically integrate partial labels across different datasets (detailed in~\S\ref{sec:method}).
It is noteworthy that the conventional classification must have a predefined and fixed number of categories, but our class query trained with a question-answer manner can handle arbitrary, varying categories, thus becoming more suitable for multiple datasets with partial labels.
Furthermore, pseudo labels and consistency constraints are introduced for the missing part of labels and for mitigating the domain gap across different datasets (see~\figureautorefname~\ref{fig:framework_overview}B).

We validate the effectiveness of Label-Assemble in both computer-aided disease diagnosis and detection, supported by two clinical applications. 
\textbf{(I)} Assembling existing labels from ChestXray14 (available since 2017) significantly improves the accuracy of COVID-19 diagnosis from 96.3\% (previous state of the art~\cite{wang2020covid}) to 99.3\%. The experiments show that assembling \textit{pathologically-related} labels can improve the diagnosis accuracy of the interested disease.
\textbf{(II)} Assembling partial labels can also help disease detection, \eg the detection of pancreatic ductal adenocarcinoma (PDAC) can greatly benefit from leveraging the labels of Cysts and PanNets (two other types of pancreatic abnormalities), increasing sensitivity from 52.1\% (previous state of the art~\cite{xia2022felix}) to 84.0\% and maintaining a high specificity of 98.0\%. The experiments also verify that assembling \textit{spatially-related} labels can help detect the interested disease more precisely.

In summary, the improved results from Label-Assemble are attributable to our simple yet powerful observation: \emph{learning from the classes of ``negative examples'' can better delimit the decision boundary of the class of interest}.
This observation agrees with the concept of ``Near Misses''~\cite{winston1976psychology,gurevich2006active}, which proposed to construct negative examples near the decision boundary to facilitate the learning of visual recognizers.
These results also suggest that rather than chasing for labels of the interested class, assembling labels of alternative classes can also lead to a substantial performance gain, especially for the minority class, \eg rare and novel diseases.
To our best knowledge, this study is among the first to systematically examine the rationale of assembling multiple datasets and fully exploit the potential of partial labels---the latest attempts~\cite{zhang2021dodnet,shi2021marginal,fang2020multi} built models on the labeled part of the data only. 

\section{Label-Assemble}
\label{sec:label_assemble}

\subsection{Motivation}
\label{sec:motivation}

We hypothesize that 
\emph{a dataset that is labeled with various classes can foster more powerful models than one that is only labeled with the class of interest.} 
To validate this point, we use six multi-class datasets. For comparison, we train a multi-class classifier and a binary classifier, wherein the interested class is labeled as positive, and the rest classes are negatives. The goal is to classify the interested class. Note that the total numbers of images are the same for training the two classifiers---the only variation is that the makeup of negatives is unknown in the binary classifier, yet it is known in the multi-class classifier. Let ``melanoma'', ``distal convoluted tubule'', ``zero'', and ``cat'' be the interested classes in the DermaMNIST, TissueMNIST, MNIST, and CIFAR10 datasets, respectively.
In the ChestXray and CheXpert datasets, five common chest diseases, \ie ``cardiomegaly'', ``pneumonia'', ``atelectasis'', ``edema'', ``effusion'', are the interested classes.
\figureautorefname~\ref{fig:framework_overview}A shows that in all six datasets, the multi-class classifier consistently outperforms the binary classifier in identifying the interested classes.
We attribute the deficient performance of the binary classifier to the lack of fine-grained labels in negative examples.
Now, we have reached a conclusion that 
\emph{learning from the classes of ``negative examples'' can better delimit the decision boundary of the class of interest}.
This conclusion has the potential to accelerate the development circle of computer-aided diagnosis and detection of novel diseases (\eg COVID-19 in late 2019), whose positive label is hard to collect, yet negative labels are usually available and relatively easier to assemble.
Normally, one would not consider using extra labels that seem unrelated to the interested class, but we find that those existing datasets, even if they were not created for the novel disease, are helpful for improving the performance and reducing annotation efforts (\figureautorefname~\ref{fig:framework_overview}C).
This has motivated the initiative of Label-Assemble, underlining the necessity of combining multiple datasets with diverse (yet partial) labels.

\subsection{Methodology}
\label{sec:method}

\textbf{\textit{Dynamic adapter with learnable class queries.}} The strategy of set prediction was initially proposed for question-and-answer tasks in NLP and has recently demonstrated its power in vision tasks, such as object detection (\eg DETR~\cite{carion2020end}), semantic segmentation (\eg MaskFormer~\cite{cheng2021per}), and medical imaging (\eg DoDNet~\cite{zhang2021dodnet}.
In light of its flexibility and effectiveness, we leverage this training strategy to address the partial label problem for the tasks of disease diagnosis and detection.
Specifically, we introduce class queries, which are initialized as one-hot vectors of each class and are \textit{learnable} during the training (differ from \cite{zhang2021dodnet}).
The class query is converted into a tensor with the same dimension as image features using a single \textit{fc} layer.
Then a prediction head, technically a linear classification layer, can generate the predictions by integrating features and class queries via inner product operations.
As shown in \figureautorefname~\ref{fig:framework_overview}B, given class queries ($q$) and input image ($x$), our dynamic adapter can compute the output ($a$) as
$a=w(q;\theta_w)*f(x)$, where $*$ is the inner product operation, $w$ is the fully connected layer transforming class queries to classification parameters, and $f$ is the feature extractor (CNN). 
Subsequently, binary cross entropy loss is used if the label ($y$) is provided, \ie
$\mathcal{L}_{\text{bce}} = -(y \cdot log(a) + (1-y) \cdot log(1-a))$.

\smallskip\noindent\textbf{\textit{Pseudo labels \& consistency constraints.}}
To unleash the full potential of unannotated labels, we introduce a sharpening operator to generate pseudo-labels, \ie
\begin{equation}
    \tilde{a}=
\begin{cases}
a+(1-a)/t, & a>{\tau}\\
a-a/t, & a\leq{\tau} 
\end{cases}
\label{eq:sharpen}
\end{equation}
where $ \tilde{a} $ is the pseudo-label of the answer, $t$ is the sharpen temperature, and $\tau$ is the threshold ($\tau=0.5$ in our experiments). The prediction beyond (below) the threshold $\tau$ can be assigned to a higher (lower) score controlled by $t$. 
If $t=\infty$, there is no pseudo-labeling; 
if $t=1$, the model converts a soft label to a completely hard label (either 1 or 0, equivalent to FixMatch~\cite{sohn2020fixmatch}).
With the sharpening operator, the loss enables the model to operate self-training on unlabeled data, \ie
$\mathcal{L}_{\text{pseudo}}= \left \| a_w - \tilde{a}_w \right \|^{2}_{2}$, where $a_w$ and $\tilde{a}_2$ denote the answer of weakly augmented images and its sharpened pseudo-labels, respectively.
To reduce the domain gap across the heterogeneous data sources, we further employ consistency constraints on weakly augmented ($a_w$) and strongly augmented ($a_s$) images. The consistency loss can be formulated as
$\mathcal{L}_{\text{consist}}=\left \| a_s - \tilde{a}_w \right \|^{2}_{2}$.

\smallskip\noindent\textbf{\textit{Overall loss function.}}
The overall loss function consists of binary cross-entropy regularization for annotated labels as well as pseudo labels \& consistency constraints for unlabeled ones, \ie
$\mathcal{L}_{\text{total}}=\mathcal{L}_{\text{bce}} + \mathcal{L}_{\text{pseudo}} + \mathcal{L}_{\text{consist}}$. Note that $\mathcal{L}_{\text{pseudo}}$ and $\mathcal{L}_{\text{consist}}$ are computed after a few warm-up epochs when the model predictions become fairly stable.

\section{Experiment, Result, and Discussion}
\label{sec:experiment}

\noindent\textbf{\textit{Dataset \& metric.}}
We evaluate our method on two computer vision datasets (\ie MNIST, CIFAR10), seven public medical datasets (\ie COVIDx CXR-2~\cite{wang2020covid}, CheXpert~\cite{irvin2019chexpert}, ChestX-ray14~\cite{wang2017chestx}, DermaMNIST, TissueMNIST, OrganAMNIST, RetinaMNIST~\cite{yang2021medmnist}), and two private medical datasets (\ie JHH and HMS)~\cite{xia2022felix}. 
Following prior metrics for benchmarking, we evaluate the performance using Area Under the Curve (AUC) for disease diagnosis; sensitivity and specificity for disease detection.
All experiments are performed by a statistical analysis based on an independent two-sample \textit{t}-test. 

\smallskip\noindent\textbf{\textit{Baseline \& implementation.}}
We compare our method with \textit{three} types of baselines: 1) the multi-network strategy~\cite{lu2021taskology} (\textit{one-model-one-task}), 2) multi-source learning algorithms \cite{chen2019med3d,zhang2021dodnet}, and 3) SOTA algorithm \cite{Kim_2021_CVPR,hermoza2020region,ma2020multilabel,taslimi2022swinchex,xiao2022delving} on NIH ChestX-ray14 and Standard CheXpert.
For a fair comparison, we choose DenseNet121 as the backbone. 
All experiments run 64 epochs and utilize Adam optimizer with an initial learning rate of 2$e$-4. We reduce the learning rate by a factor of 2.0 on the plateau with 5 steps of patience. Early stopping patience is set to be 10 epochs. The pseudo-label threshold $\tau$ and the sharpen temperature $t$ are 0.5 and 4.0, respectively. 

\begin{table*}[t]
\centering
\footnotesize
    \caption{Assembling 75,310 \textit{partial} labels, our method outperforms other methods developed for partial labels, and performs on par with the method learning from 105,434 \textit{full} labels, eliminating the need for additional 40\% annotation costs.
    The performance is measured by AUC. No significant difference ($p>0.05$) between ours (75K partial labels) and DenseNet (105K full labels).}
    \begin{tabular}{p{0.09\linewidth}P{0.05\linewidth}|P{0.04\linewidth}P{0.04\linewidth}P{0.04\linewidth}P{0.04\linewidth}P{0.05\linewidth}P{0.05\linewidth}|P{0.04\linewidth}P{0.04\linewidth}P{0.04\linewidth}P{0.04\linewidth}P{0.05\linewidth}P{0.05\linewidth}}
    \hline
    &  & \multicolumn{6}{c|}{CheXpert (val)} & \multicolumn{6}{c}{ChestX-ray14 (val)}\\
    Method & \# labels & Card$^{\dagger}$ & Pneu1$^{\dagger}$ & Atel$^{\dagger}$ & Edema & Effusion & Average & Cons$^{\dagger}$ & Pneu2$^{\dagger}$ & Atel$^{\dagger}$ & Edema & Effusion & Average\\
    \shline
     DenseNet~\cite{lu2021taskology} & 37,655 & 0.646 & 0.461 & 0.431 & 0.791 & 0.800 & 0.626 & 0.693 & 0.640 & 0.688 & 0.737 & 0.783 & 0.708  \\
     Med3D~\cite{chen2019med3d} & 75,310  & 0.751 & 0.629 & 0.663 & 0.839 & 0.836 & 0.744 & 0.700 & 0.758 & 0.718 & 0.732 & 0.788 & 0.739 \\
     DoDNet~\cite{zhang2021dodnet} & 75,310 & 0.778 & 0.598 & 0.646 & 0.859 & 0.845 & 0.745 & 0.706 & 0.756 & 0.721 & 0.745 & 0.769 & 0.740\\
     Ours & 75,310 & 0.832 & 0.675 & \textbf{0.702} & \textbf{0.867} & \textbf{0.886} & 0.792 & \textbf{0.744} & 0.805 & \textbf{0.813} & 0.710 & 0.778 & 0.770\\
     \hline
     DenseNet~\cite{lu2021taskology} & 105,434 & \textbf{0.835} & \textbf{0.683} & 0.699 & 0.864 & 0.885 & \textbf{0.793} & 0.719 & \textbf{0.810} & 0.740 & \textbf{0.811} & \textbf{0.812} & \textbf{0.778} \\
    \hline
    \end{tabular}
    \begin{tablenotes}
    \item $^{\dagger}$Card, Pneu1, Atel, Cons, Pneu2 denote Cardiomegaly, Pneumonia, Atelectasis, Consolidation Pneumothorax, respectively.
    \end{tablenotes}
    \label{tab:comparison_partial_full_labels}
\end{table*}

\subsection{Assembling partial labels improves disease \textit{diagnosis}}
\label{sec:improved_diagnosis}

As shown in~\S\ref{sec:motivation}, learning with additional ``negative examples'' improves the performance of the interested class(es).
For example, assembling existing labels from ChestXray14 significantly improves the accuracy of COVID-19 diagnosis from 96.3\% to 99.3\% and improves the AUC of nodule diagnosis from 0.69 to 0.78.
However, how much different classes of ``negative examples'' contribute to the performance remains unknown.
We further delve into this problem and find that \emph{the performance gain is positively related to the pathological similarity between the interested class and the added classes.} 
\figureautorefname~\ref{fig:framework_overview}C illustrates the improvements of classifying ``Nodule'' by assembling images of 13 different diseases. 
The Pearson correlation coefficient between the similarity\footnote{The similarity is quantified by the Cosine distance between the two learned class queries (see~\figureautorefname~\ref{fig:framework_overview}B and \S\ref{sec:method}).} and the performance gain is 0.83, which indicates a significant positive correlation ($p=4.93e$-$4$). 
This means that \textit{assembling pathologically similar classes is more beneficial than dissimilar classes for the interested class}.
A similar observation is obtained in the example of COVID-19 diagnosis (see gray bars in~\figureautorefname~\ref{fig:framework_overview}C).
In practice, it is hard to obtain enough labels for training since novel diseases have limited positive examples. 
By assembling similar diseases from other publicly available datasets, the model can better identify novel diseases, thus relieving the long-tail problem in computer-aided diagnosis.
Interestingly, pseudo labels and consistency constraints can largely eliminate the requirement of similar diseases, suggesting that assembling any chest disease (regardless of the specific classes) can achieve equally high performance of COVID-19 diagnosis (see red bars in~\figureautorefname~\ref{fig:framework_overview}C). These results are encouraging, but more investigation will be needed.

\subsection{Assembling partial labels improves disease \textit{detection}}
\label{sec:improved:detection}

JHH and HMS datasets are used to detect PDAC from CT scans.
The detection of PDACs can be influenced by other types of pancreatic abnormalities, \eg pancreatic cysts and Pancreatic Neuroendocrine Tumors (PanNETs), regarding their appearance, intensity, texture, and so on. 
With the same number of training cases (1,195) from JHH, we train two models: the first one only segments PDACs from the background, and the second one is trained to segment all three types of tumors from the background.
These two models are evaluated on the JHH test set and HMS on the performance of detecting PDACs. As shown in~\figureautorefname~\ref{fig:framework_overview}A, the performance of PDAC detection in JHH test set increases from 52.1\% to 84.0\% by exploiting labels of Cysts and PanNETs, while maintaining a high specificity of 98.0\%; the performance of PDAC detection in HMS test set increases from 81.5\% to 93.5\%, while maintaining a high specificity of 90.2\%.

\subsection{Combining partial labels vs. full labels}
\label{sec:on_par_full_label}

We also compare with the methods \cite{chen2019med3d,zhang2021dodnet} developed for partial labels.
Med3D~\cite{chen2019med3d} adopts the multi-network strategy (one-model-one-task) and DoDNet~\cite{zhang2021dodnet} learns multiple tasks in one network with shared feature extractor.
Our method differs from them in \textit{two} perspectives: (1) our adapter with updated encodings enables the model to capture the relations of classes and benefits multi-label learning, and (2) we use pseudo-labeling and consistency loss to exploit unannotated data.
To adapt their method to our setting, we utilize 15,062 images from ChestX-ray14 and CheXpert with seven diseases labeled and three out of the seven diseases are shared between the two datasets.
The results in \tableautorefname~\ref{tab:comparison_partial_full_labels} indicate that \textbf{(I)} our method with the adapter and semi-supervised learning framework achieves a better performance of multi-task learning, and \textbf{(II)} our method enables learning from partial labels to perform on par with that from full labels while eliminating an additional annotation cost of 40\% (75,310 partial labels~vs.~105,434 full labels).
The obtained results indicate that 
\emph{it is not necessary to complete the missing labels in an assembly of multiple partially labeled datasets.} 

\subsection{Exceeding Prior Arts in NIH ChestX-ray14}
\label{sec:sota}

\begin{table}[!h]
\footnotesize
\centering
    \centering
    \caption{
        Label-Assemble achieves the best mean performance over all 14 thorax diseases on ChestXray-14 (\emph{official} split).
    }
    \begin{tabular}{p{0.27\linewidth}p{0.23\linewidth}p{0.2\linewidth}P{0.1\linewidth}}
    \hline
         & Ref.~\&~Year & Architecture & mAUC \\
        \shline
        Ma~\etal~\cite{ma2020multilabel} & MICCAI~2019 & DenseNet ($\times$2) & 0.817\\
        Hermoza~\etal~\cite{hermoza2020region} & MICCAI~2020 & DenseNet121 & 0.821\\
        Kim~\etal~\cite{Kim_2021_CVPR} & CVPR~2021 & DenseNet121 & 0.822\\
        Taslimi~\etal~\cite{taslimi2022swinchex} & arXiv~2022 & SwinT & 0.810 \\
        Xiao~\etal~\cite{xiao2022delving} & WACV~2022 & ViT-S & 0.823 \\
        \hline
        Ours & & DenseNet121 & \textbf{0.832}\\
    \hline
    \end{tabular}
    \label{tab:sota_chestxray}
\end{table}
Table~\ref{tab:sota_chestxray} shows that assembling partial labels from publicly available datasets sets a new state of the art on ChestX-ray14 (mAUC = 0.832), yielding the best performance on 13 out of 14 diseases. Similarly, Label-Assemble is also effective on the CheXpert dataset with a 1.8\% improvement over the baseline.

\section{Conclusion}
\label{sec:conclusion}

We propose a new initiative, Label-Assemble, to explore the full potential of an assembly of publicly available datasets with partial labels.
The rationale of the initiative is validated on a total of six medical datasets, showing that assembling pathologically-related and spatially-related labels are preferred for disease diagnosis and detection, respectively.
This is particularly valuable for novel disease diagnosis, underlining the role of an assembly of existing labels of related diseases, rather than narrowly pursuing expensive labels for the interested class.
This work represents the foremost step towards creating large-scale, multi-center, fully-labeled medical datasets---one of the foundations of fostering future research in deep learning applied to medical images.

\clearpage

\noindent\textbf{Compliance with Ethical Standards.}
Ethics committee/IRB of Johns Hopkins Medicine gave ethical approval for this work.

\smallskip\noindent\textbf{Acknowledgments.}
This work was supported by the Lustgarten Foundation for Pancreatic Cancer Research and the McGovern Foundation.
We thank M.~R.~Hosseinzadeh~Taher and F.~Haghighi for surveying top solutions on NIH ChestX-ray14 (\tableautorefname~\ref{tab:sota_chestxray}); thank R.~Feng for reproducing baseline methods on NIH ChestX-ray14.
We also thank J.~Liang, Z.~Zhu, L.~Chen, J.~Chen, Y.~Bai, G.~Li, and X.~Li for the constructive suggestions; O.~Welsh for improving the writing of this paper.

{\small
\bibliographystyle{IEEEbib}
\bibliography{refs}
}

\end{document}